# Face Synthesis (FASY) System for Determining the Characteristics of a Face Image

Santanu Halder[1], Debotosh Bhattacharjee[2], Mita Nasipuri[2], Dipak Kumar Basu[2*], Mahantapas Kundu[2]

[1]Department of Computer Science and Engineering, GNIT, Kolkata - 700114, India,
Email: sant.halder@gmail.com

[2] Department of Computer Science and Engineering, Jadavpur University, Kolkata, 700032, India  [*]AICTE Emeritus Fellow

**ABSTRACT**

This paper aims at determining the characteristics of a face image by extracting its components. The FASY (FAce SYnthesis) System is a Face Database Retrieval and new Face generation System that is under development. One of its main features is the generation of the requested face when it is not found in the existing database, which allows a continuous growing of the database also. To generate the new face image, we need to store the face components in the database. So we have designed a new technique to extract the face components by a sophisticated method. After extraction of the facial feature points we have analyzed the components to determine their characteristics. After extraction and analysis we have stored the components along with their characteristics into the face database for later use during the face construction.

## 1. INTRODUCTION

Face is most important visual identity of a person and while meeting an unknown person, it is the face that attracts our attention most. We often describe a person in terms of the characteristic features of important face components like eyes, eyebrows, nose and lip together with the overall shape of the face. The criminal Investigation agencies employ artists to sketch human faces from the description given by a witness about criminal's appearance which is then searched in the face database of the known criminals for recognition. Since early 1990's, Face Recognition Technology (FRT) becomes an active research area and there are a lot of works on facial recognition and facial feature extraction [1] [2]. This paper presents a much simpler, vivid and easy to comprehend method for extraction of feature points and thereby, analyzing the characteristics of human facial components. Also the method is computationally inexpensive unlike many of the other methods available for this purpose. This paper is a part of a main research effort, whose aim is the construction of the FASY-System based on human like description of each components of the face. The FASY System has the following features: (i) Extract the components from a face image (ii) Analyze the extracted components and store the information along with the component into database for later use, (iii) face queries using human-like description of the face, (iv) searching the required face in the database and (v) generation of the requested face with the stored components when they are not in the database. We have described the last three features in our paper titled "Face Synthesis (FASY) System for Generation of a Face Image from Human Description" which has been accepted in the third International Conference on Industrial and Information Systems (ICIIS) that will be held at Kharagpur, INDIA from December 8-10, 2008. This paper is focused on first two features. The first and the most important step in feature detection is to track the position of the eyes. Thereafter, the symmetry property of the face with respect to the eyes is used for tracking rest of the features like eyebrows, lips and nose. Splitting face into two halves eases the process further.

## 2. PREPROCESSING

Here we have worked with our own r-g-b face database. Generally, the quality and the size of images captured with mobile appliances vary with the capacity of the appliance and the users' setting. The faces are cropped from hair to chin (top to bottom) and from right ear to left ear (left to right). In the suggested algorithm, the size of the image has been standardized as having maximum of 200 pixels in width. When the actual size is larger than this, keeping the ratio of width and height unchanged, the image has been reduced to 200 pixels in width. During capturing images, we have tried to minimize the variation due to head tilt, shift, rotation, scaling and light effect as much as possible. Fig. 1 shows some face images and their cropped images.

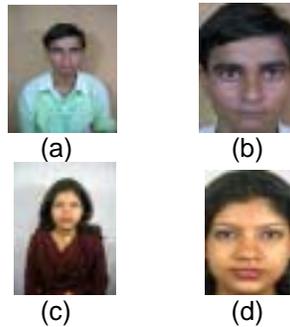

(a) (b)
(c) (d)
Fig. 1: Original face images followed by cropped images

## 3. THE FASY SYSTEM

A human user makes a query of a face using a human-like face description. The FD module interprets the query and translates it into a face description using different Face Parameters. This face description is used by the FR module to search the face in the database. The system retrieves the existing faces based on the user description. In the case when the desired face is not found in the database, the user can select the automatic generation of it. The FG module first finds the face components according to the user's requirements and generates a new face with the selected face parameters. As a result of the generation process the generated face is presented to the user. If the user is satisfied with the generated face, the process stops here and the generated face is stored in the database. Otherwise, the user enters into an iterative process. The iterative generation of the faces is implemented in the PA module. To extract the face components along with their characteristics user can give input a face image to the FCEA module which stores the components and their characteristics into the database also. These stored components are used by the FG module for finding the face components at the time of new face generation. Fig. 2 shows the block diagram of FASY system. The FCEA module has been described in the Fig. 3. FCEA contains six extractor modules and six analyzer modules for six facial components. At the end, it stores the components and their characteristics into the database.

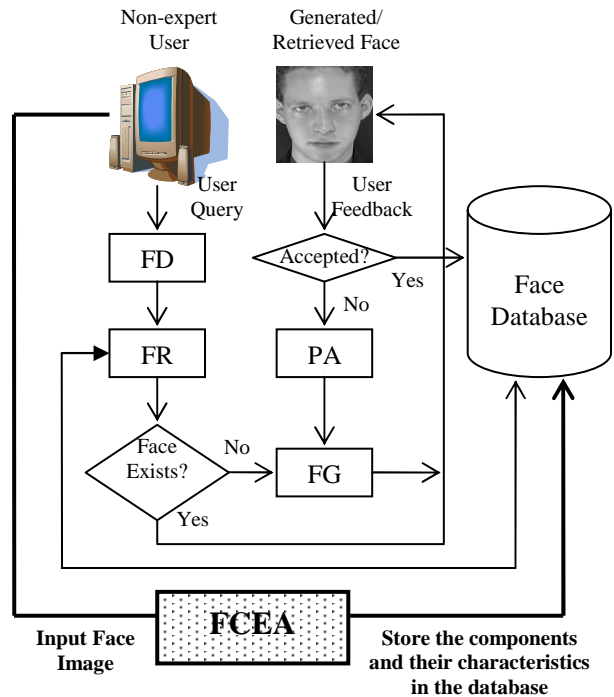

**FD**: Face Descriptors  **FR**: Face Retrieval  **FG**: Face Generation
**PA**: Parameter Adjustment  **FCEA**: Facial Components Extractor and Analyzer
Fig. 2: Block Diagram of FASY System

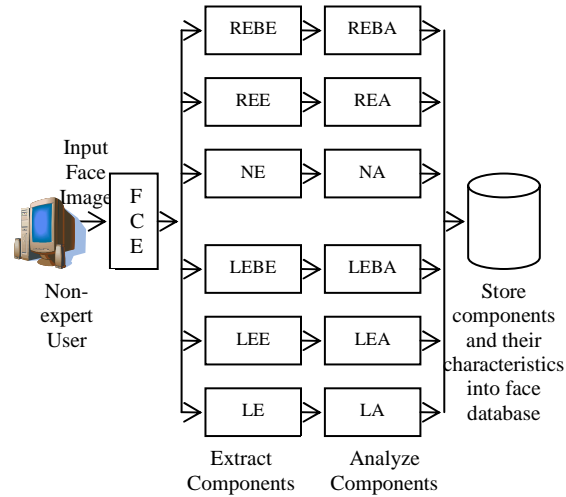

**REBE**: Right Eyebrow Extractor  **REE**: Right Eye Extractor  **NE**: Nose Extractor
**LEBE**: Left Eyebrow Extractor  **LEE**: Left Eye Extractor  **LE**: Lip Extractor
**REBA**: Right Eyebrow Analyzer  **REA**: Right Eye Analyzer  **NA**: Nose Analyzer
**LEBA**: Left Eyebrow Extractor  **LEA**: Left Eye Analyzer  **LA**: Lip Analyzer

Fig. 3: Block Diagram of Facial Components Extractor and Analyzer (FCEA) module

## 4. FACIAL COMPONENT EXTRACTION

Since the appearances of facial components are different from each other, it is not possible to extract all the facial components with one algorithm. Therefore, we prepare an algorithm optimized for each facial component. When a face image is fed into the system, we first predict the positions where the facial components should appear at, and then we detect the facial components by applying the proper algorithm to the area where a facial feature is predicted to appear at. For this prediction of the component regions we have performed all the calculations in respect of the width of the face (W). Suppose we are extracting the components for the face of fig. 1(d).

### 4.1 Right Eye Extraction

We are assuming that the right eye lies in the right half of the face. Now maximum difference of intensities between two adjacent pixels occurs in the middle of the eye. Therefore if we get the row on which the maximum difference occurs then we can predict the eye region. The steps to extract the right eye are as follows:

1. Extract the right half of the face (RFace).
2. Find the gray image (RFaceGray) of Rface.
3. Find the row (Max_Index) on which the maximum difference between two adjacent intensities in RFaceGray lies.
4. Now extract a portion of RFaceGray according to the following calculation.

(X1,Y1)
REye
(X2,Y2)

$X1 = Max\_Index - (0.06*W)$
$X2 = Max\_Index + (0.06*W)$
$Y1 = 0.14*W$
$Y2 = W/2$

5. Apply median filter on Reye.
6. Normalize REye by adding 127 with REye and get a new matrix REyeNormalized.
7. Find edges in REyeNormalized by canny's edge detection algorithm which yields a new matrix REyeNormalizedCanny.
8. Using a boundary detection algorithm, get the exact right eye region from REyeNormalizedCanny.

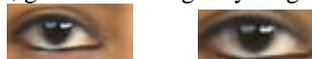

(a)           (b)

Fig. 4: (a) Predicted right eye region
(b) Exact right eye region after applying the method

## 4.2 Nose Extraction

1. Find the gray image (FaceGray) of the face.
2. Predict the nose region as the following calculation:

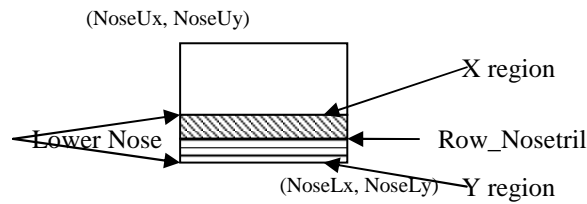

$X1 = Max\_Index - (0.15*W)$
$X2 = X1 + (0.55*W)$
$Y1 = 0.32*W$
$Y2 = Y1 + 0.4*W$

3. Get the upper boundary from the binary image of Nose.
4. Extract the lower $1/3^{rd}$ portion (LowerNose) of the Nose.
5. Normalize the LowerNose by adding 127 with it and get a new matrix LowerNoseNormalized.
6. Find the row (Row_Nostrils) on which nostrils exists (by getting the minimum intensity in LowerNoseNormalized).
7. Find the edges of LowerNoseNormalized using canny's edge detection algorithm and get the logical matrix LowerNoseNormalizedCanny.
8. Now the left and right boundary of nose lies in the X region and lower boundary lies in the Y region of the Fig. 5. Therefore from the LowerNoseNormalizedCanny get the left, right and lower boundary of nose.

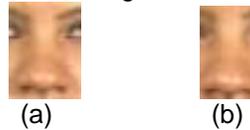

Fig. 5

(a)           (b)

Fig. 6: (a) Predicted nose region
(b) Exact nose region after applying the method

## 4.3 Right Eyebrow Extraction

For extraction of the right eyebrow, first we predict the position where the right eyebrow can exist. Then applying the method described in [3] on the predicted region, we extract the exact eyebrow region.

1. Predict the right eyebrow region by the following method:

$X1 = Max\_Index - (0.15*W)$
$X2 = Max\_Index$
$Y1 = 0.05*W$
$Y2 = W/2$

2. Convert REyebrow from r-g-b scale to HSV scale.
3. The following steps are performed to convert the V values to a binary scale:
    (a) Let P=0.25  Q=0.25
    (b) if $|V-P| > Q$ then T=0 else T=1
    (c) if T=1 then $F = 1-((V-P)/Q^2)$ else F=0
    (d) if F<0.5 then C=1 else C=0

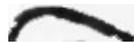

Fig. 7: Extracted Right eyebrow in HSV scale using V values

## 4.4 Lip Extraction
1. Predict the lip region as:

```
(X1,Y1)
   ┌─────────┐        X1 = NoseLx + (0.05*W)
   │   Lip   │        X2 = NoseLx + (0.30*W)
   └─────────┘        Y1 = 0.25*W
         (X2,Y2)      Y2 = Y1 + 0.5*W
```

2. Apply median filter on gray image of Lip (LipGray).
3. Normalize Lip by adding 127 (LipGrayNormalized).
4. Find edges of LipGrayNormalized using canny's edge detection algorithm and get the new logical matrix LipGrayNormalizedCanny.
5. Using a boundary detection algorithm, find the boundary of the lip from LipGrayNormalizedCanny.

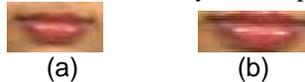

(a)        (b)

Fig. 8: (a) Predicted lip region
(b) Exact lip region after applying the method

## 4.5 Left Eye and Left Eyebrow Extraction
We are assuming that the left eye lies in the left half of the face. Now like as the right eye, maximum difference of intensities between two adjacent pixels occurs in the middle of the left eye. Left eye extraction method is as same as right eye extraction except the prediction on the region. The calculation for the prediction of the left eye region is:

```
(X1,Y1)
   ┌─────────┐        X1 = Max_Index – (0.06*W)
   │  LEye   │        X2 = Max_Index + (0.06*W)
   └─────────┘        Y1 = W/2
         (X2,Y2)      Y2 = W - (0.05*W)
```

Similarly the extraction method of left eyebrow is as same as the extraction of right eyebrow except the prediction of the left eyebrow region and the region is calculated as:

```
(X1,Y1)
   ┌─────────┐        X1 = Max_Index - (0.15*W)
   │ LEyebrow│        X2 = Max_Index
   └─────────┘        Y1 = W/2
         (X2,Y2)      Y2 = W - (0.05*W)
```

## 5. FACIAL COMPONENT ANALYSIS
Here we have analyzed each component to determine the size of them in terms of their width and height. After extraction of each components their width and size is easily computable from the upper left and bottom right co-ordinates of them. Suppose after extraction, the width and height of a component is CW and CH respectively and its normal width and height in respect of the face width (W) are W*fw and W*fh respectively where fw and fh are two fractions lie between 0 and 1 ($0 \leq fw, fh \leq 1$). Now the width and height are calculated for the component is calculated in percentage as follows:

Width = 50 – (W*fw – CW) * 50
Height = 50 – (W*fh – CH) * 50

Now set the two limits x and y such that if the percentage lies between 0 to x-1 then the component is small, if the percentage lies between x to y then the component is normal and otherwise large. From experimental result we found that the values of fw and fh for the components are as follows:

fw for eye = 0.24 and fh for eye = 0.12
fw for nose = 0.27 and fh for nose = 0.47
fw for lip = 0.40 and fh for eye = 0.13

## 6. EXPERIMENTAL RESULTS

For testing the proposed method, we had about 200 male and female face images of different ages. The following snapshots show some of the results where face components have been extracted successfully.

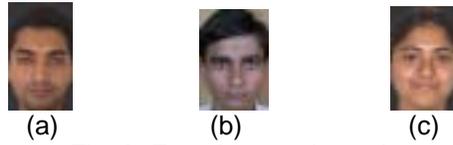

(a)         (b)         (c)
Fig. 9: Face Images in r-g-b

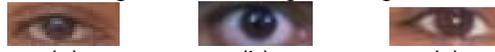

(a)         (b)         (c)
Fig. 10: Extractions of right eyes in r-g-b

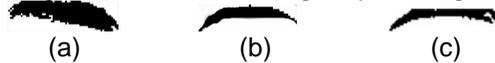

(a)         (b)         (c)
Fig. 11: Extraction of Right eyebrow in HSV scale using the V values

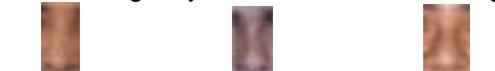

(a)         (b)         (c)
Fig. 12: Extraction of the nose in r-g-b

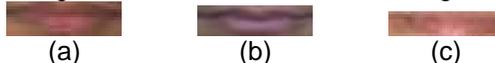

(a)         (b)         (c)
Fig. 13: Extraction of the lips in r-g-b

Now the analysis for the extracted components for the face images given in Fig. 9 is described in the TABLE 1.

| Face | Component | Width | Height |
|---|---|---|---|
| Fig 9(a) | Eye | 54.16 | 41.66 |
| | Nose | 51.85 | 55.31 |
| | Lip | 49.37 | 50.00 |
| Fig. 9(b) | Eye | 48.61 | 63.88 |
| | Nose | 43.75 | 48.57 |
| | Lip | 44.16 | 34.21 |
| Fig. 9(c) | Eye | 52.08 | 41.66 |
| | Nose | 53.7 | 48.40 |
| | Lip | 51.87 | 51.92 |

TABLE 1: Analysis of the components

## 7. CONCLUSION

This paper described the construction of a face database and presented some analysis of the images that were collected as a part of this project. When the user extracts the face components, the database can be expanded and the characteristics of the extracted face components are stored in the database as well. From this, even a small-at-the-beginning face database can grow from the very beginning on.